# A Generalized Method for Integrating Rule-based Knowledge into Inductive Methods Through Virtual Sample Creation

**Ridwan Al Iqbal**

American International University-Bangladesh

Dhaka, Bangladesh

stopofeger@yahoo.com

## Abstract

Hybrid learning methods use theoretical knowledge of a domain and a set of classified examples to develop a method for classification. Methods that use domain knowledge have been shown to perform better than inductive learners. However, there is no general method to include domain knowledge into all inductive learning algorithms as all hybrid methods are highly specialized for a particular algorithm. We present an algorithm that will take domain knowledge in the form of propositional rules, generate artificial examples from the rules and also remove instances likely to be flawed. This enriched dataset then can be used by any learning algorithm. Experimental results of different scenarios are shown that demonstrate this method to be more effective than simple inductive learning.

## 1 Introduction

The majority of machine learning algorithms learn from scratch, training models from examples without considering existing domain knowledge. Unlike human learners, these systems are not capable of accumulating domain knowledge and sequentially improving their ability to learn tasks. These algorithms require significant amount of training data and training time to perform well [Kearns & Vazirani, 1994]. Domain knowledge in learning has been shown to boost learning speed and accuracy significantly [Pazzani et al., 1997]. This is due to the fact that hybrid systems use information from one source to offset for the misinformation in the other source. This is necessary as real world learning problems are complex enough that a considerable amount of training data is required; but training data is quite expensive to acquire and also prone to noise. This lack of data can be compensated by domain knowledge which is equally expensive and imperfect. Thus a balanced system that uses both always performs better.

There have been many such hybrid systems that use different forms of domain knowledge. These include propositional rules such as KBANN [Towell & Shavlik, 1994], FOCL [Pazzani et al., 1991]; certainty factors such as RAPTURE [Mahoney & Mooney, 1992]; Invariance [Schölkopf et al., 1996]; hints [AbuMostafa, 1995] and so on. Ting Yu's PhD dissertation [Yu, 2007] provides a systematic overview of hybrid machine learning algorithms. Yu identified four categories or methods of providing knowledge in hybrid methods. These methods also follow the steps taken by typical inductive learning algorithms. These are: 1) Using prior knowledge to preprocess training data; 2) using prior knowledge to initiate the hypothesis; 3) using prior knowledge to alter the search objective; 4) using prior knowledge to augment search operation.

Only the first method; that is using prior knowledge to preprocess training data is not dependent on the underlying machine learning algorithm. This method tailors the training data by cleaning, selecting and augmenting. The scarcity of training data and noise in training data is the major concern of inductive algorithms. Domain knowledge preprocessors can remove flawed training examples as well add new virtual training examples. Most popular machine learning algorithms are dependent on training dataset alone. Virtual training examples offer a general purpose way of including domain knowledge that can be used by any inductive algorithm.

This method of adding virtual examples has gained popularity in the recent years. Niyogi used virtual examples to add Invariance in Neural Networks [Niyogi et al., 1998], Schölkopf [Schölkopf et al., 1996], Decoste [Decoste & Schölkopf, 2002], Sassano [Sassano, 2003] all researched the use of virtual examples to introduce Invariance in Support vector machines. However, these methods have been used predominantly to introduce Invariance. Invariance is the knowledge that certain transformations of a sample are equivalent with each other. Thus invariant learners are unaffected by perturbation in the data.

A predominant form of domain knowledge is propositional rules or production rules. Human experts can express their knowledge in terms of rules more easily. Therefore, having methods that can take domain knowledge in the form rules is would be a great improvement.

However, no such method exists that can preprocess training data and can be used by any learning algorithm.

Our research introduces a method called RASCAL (Rule Assisted Sample Creation And Deletion) that uses propositional rules to preprocess datasets. This method adds virtual examples that conform to the provided rule set. This is a hybrid method that does not consider either rule set or training dataset to be error-free. The rules in the rule set are judged by a scoring criterion that includes accuracy, range and generality. Training samples are also judged by their conformance to the rules and samples considered flawed are pruned from the dataset. This way both source can compensate for each other's deficiencies. The resultant dataset incorporates the domain knowledge. Then the refined dataset can be used by any algorithm. This universality makes RASCAL highly robust.

In this paper we explain RASCAL and analyze the performance when applied to real world problems. The real world domain of molecular biology specifically the problem of recognizing *eukaryotic splice junctions*, *promoter genes* is used for the experiments. It is shown that RASCAL refined datasets perform much better than the original dataset even if less data is provided.

Section 2 describes the algorithm while section 3 shows the experimental results. Section 5 concludes the paper.

The common used notation is as follows: N, size of dataset. K, Number of features with A being the feature set. R is the ruleset with $R_o$ being the operational ruleset.

## 2  Description of the RASCAL algorithm

The goal of RASCAL is to have a dataset that conforms to the rules. The motivation behind this is straightforward. The rules embody domain knowledge and must surely have some validity. Virtual examples that conform to the rules can be generated simply by having the samples follow the conditions and class value of the rules. However, simply following this would degrade performance if the rule set is misleading. This is why RASCAL follows a method of check and balance.

The RASCAL algorithm works in three steps. First, the provided rule set is preprocessed and normalized to conform to the specification required by the algorithm. Then the normalized rules are evaluated against the data set and a score is calculated. Finally, virtual training samples are generated based on the score while examples that are likely to be error-prone are also removed.

### 2.1  Preprocessing Rules

RASCAL only supports rules in the form of antecedents followed by consequent. Thus, it is of the following form:

If *Precondition* THEN *Postcondition*

The antecedents can be a mixture of conjunction and disjunction of *literals*. So, the following is a valid rule set:

**Example Ruleset 1:**

$x \land \neg y \lor z \vdash k$

$\neg a \land \neg b \vdash j$

$\neg j \land k \vdash Class$

In the example ruleset above, the literals j and k are implicated by other literals while j and k refers to the actual class. Therefore, a hierarchy of rules can be made where each successive layer of literals are made up from other literals. However, the algorithm cannot work with rules in this form. We must first draw a distinction between two kinds of literals that appear in the rule set. We will say a literal is atomic if it is not made up of other literals. Thus only input features of the dataset are atomic. Any other literal that is made up of other literals is called non-atomic.

The goal of preprocessing step is the conversion of all rules into *operational* form. An operational rule has the class variable as post-condition and is made up of a conjunction of atomic literals. Hence, logical OR cannot appear in operational rules. All non-operational rules can be converted into operational form.

**Definition 1.** *A rule r in the rule set R is called operational iff its antecedents are a conjunction of atomic literals and its consequent is the Class attribute.*

$r \, \epsilon \, R_o \quad iff \quad \forall x \, \epsilon \, Ant(r) \;\; x \, \epsilon \, A \, \& \, Con(r) = Class$

The preprocessing step takes all non operational rules and converts them into operational form. This is done in a backward chaining fashion. The mode of operation of this algorithm is summarized below:

1) The algorithm first selects the rules whose head (post-condition) matches the target class attribute. The preconditions of the selected rule form a logically sufficient condition for the target concept. The algorithm takes each non-atomic literal and finds the associated rule that has that particular literal as post-condition. The non-atomic literal is then substituted with its pre-conditions. These substituted literals may also contain non-atomic literals which are then again substituted by their pre-conditions. This process of "unfolding" the domain theory continues until the sufficient conditions have been restated in terms of atomic literals. For example, the sample ruleset 1 will be unfolded as:

$\neg j \land k \vdash Class$

After substitution and applying De Morgan's law becomes:

$(a \lor b) \land ((x \land \neg y) \lor z) \vdash Class$

2) After substitution the algorithms checks for the disjunctions and generates rules for each of operands of the disjunction. Thus, $A \lor B \vdash C$ becomes two rules $A \vdash C$ and $B \vdash C$. This process continues until all rules are in operational form. The example rule set is converted to operational form below.

*Substituting $a \lor b$:*

$a \land ((x \land \neg y) \lor z) \vdash Class$

$b \wedge ((x \wedge \neg y) \vee z) \vdash Class$

*Then the final operational form will be:*

**Sample Ruleset 1 (Operational) :**

$a \wedge x \wedge \neg y \vdash Class$
$a \wedge z \vdash Class$
$b \wedge x \wedge \neg y \vdash Class$
$b \wedge z \vdash Class$

As we can see, disjunctions quickly increase the number of rules. So, the input rule set should have as small number of disjunctions as possible otherwise the number of rules can quickly become intractable.

## 2.2 Score Calculation

After generating the operational rule set $R_o$, our algorithm would calculate a score that would determine the utility of the rule. This utility score would be used for both virtual example generation and the existing dataset pruning. As we consider both training set and rule set to be imperfect, so the scoring criterion cannot put more faith on one of the source. Thus our scoring function should be balanced.

The most obvious evidence of utility is accuracy or correctness; which is the ratio of the number of examples correctly classified by the rule and the number of examples that actually matches the precondition of the rule. To avoid the possibility of a division by zero, we add 1 in the divisor.

*Correctness,* $C(r) = \frac{S(r)}{M(r)+1}$ (1)

$S(r)$ = *number of examples successfully classified.*
$M(r)$ = *number of examples matched*

However, correctness is a ratio, so even rules that only match few instances can have very high correctness. For instance, a rule that only matches one example and also correctly classifies it will have a C(r) of 1 while a rule that matches 40 examples and is correct in 35 cases will have a C(r) of 0.875. However, the second rule is clearly more useful than the first. Thus correctness alone is not a sufficient criterion of utility. Instead utility should be composed of correctness and the scope or extent of the rule.

The scope E(r) of a rule defines the ratio between M(r) and the number of instances in the dataset N.

$E(r) = \frac{M(r)}{N}$ (2)

Both of these ratios are in the range of 0 to 1. We combine these two values to get the sample utility $U_s(r)$.

$U_s(r) = \frac{1}{2}C(r) + \frac{1}{2}E(r)$ (3)

Combining E(r) and C(r) this way provides a balance between both criteria. A rule that has high correctness but low scope is less useful than a rule with high scope and reasonable correctness. On the other hand, we don't want to give more utility to rules that have high scope but perform poorly on those rules. This utility function gives such reasonable measure. $U_s(r)$ is a good measure of utility in terms of sample dataset. $U_s(r)$ would be used for pruning training instances.

However, this function only measures the utility of a rule in terms of its influence and compliance with the training dataset. But when we want to generate new training instances, good generalization is the goal instead of just conformance with the dataset. There can be rules that are highly specific and the training set can be skewed to that particular rule. In such a case, $U_s(r)$ value will be high. Generating more samples from such a specific rule is not desirable as this can lead to a high concentration of training instances in a small region.

Therefore, $U_s(r)$ must be augmented by a criterion that will also measure the generalization capability of a rule. A useful observation on the rule set is that an operational rule has a precondition that is simply a conjunction of feature values. Therefore, there can be at most K literals in the precondition where K is the total number of features in dataset. Since a rule is only conjunction of literals, a rule becomes more and more specific as literals are added [Mitchell, 1997]. A rule that has all K features can have only one possible match. If we consider all features to be binary then an operational rule with j literals can have $2^{K-j}$ possible match. Therefore, a general rule is the one with fewer literals. If L(r) is the number of literals in r. then

*Generalization capability,* $G(r) = \frac{K - L(r)}{K}$ (4)

Hence, the utility of a rule based on all three criteria:

$U(r) = \frac{1}{3}C(r) + \frac{1}{3}E(r) + \frac{1}{3}G(r)$ (5)

This is the scoring function that is used to evaluate rules for virtual sample generation.

## 2.3 Sample Generation

The final step of RASCAL is to actually generate virtual examples from the rules as well as remove examples that have been deemed incorrect by the ruleset.

RASCAL takes two constants *D* and *I* as input Along with ruleset and dataset. Both of these values would greatly affect performance as they denote how many changes will take place in the dataset. *I* denotes the number of virtual samples to be generated from rules expressed as percentage of the training data set. This value will significantly influence performance of the preprocessed dataset. Too many virtual examples may hamper performance while less virtual examples would not yield noticeable performance gain. *D* is the threshold for pruning an example; thus setting a higher value will result in less pruning. *D* will be explained in 2.4.

The main goal of virtual samples is to match the rule from which they are to be created. That is, a virtual sample must have the feature values that are present in precondition of the rule. This is the necessary condition required for the sample to be a valid sample spawned from a rule. The features that are not present in the rule can be set arbitrarily as that is not the concern of the rule. However, we would prefer the virtual examples to be distributed as randomly as possible; as this would increase the probability of a good generalization.

The virtual sample generation of RASCAL works by generating an initial template for the examples to be generated from a Rule r. The preconditions in the rule become the value of those features. Finally the attributes missing in the rule are set randomly. The class label is taken from the rule. This creates a valid example. The sample generation of a rule taken from the earlier generated operational *Ruleset 1* is shown below:

$$a \wedge x \wedge \neg y \vdash Class$$

A complete instance is made of 5 features a, b, x, y, z. There are 3 literals present in the rule. Thus the template for sample generation will be:

$$a = T, b =??, x = T, y = F, z =?? \; Class = T$$

Now we generate 2 instances by randomly selecting the values for b and z.

$$a = T, \mathbf{b} = \mathbf{RAND()} = \mathbf{T}, x = T, y = F, \mathbf{z} = \mathbf{RAND()} = \mathbf{F} \; Class = T$$

$$a = T, \mathbf{b} = \mathbf{RAND()} = \mathbf{F}, x = T, y = F, \mathbf{z} = \mathbf{RAND()} = \mathbf{F} \; Class = T$$

This is how Virtual samples are generated for each rule. The algorithm now just needs to calculate the number of virtual samples to be generated from each rule P(r) and use the method described above to generate P(r) instances.

The score U(r) generated in the earlier step is used to calculate P(r). We want higher scoring to generate more instances however low scoring rules should not be totally left out either. So, we normalize the scores of each rule with the total score of all rules.

$$Total\ score = \sum_{i \in R} U(i)$$

$$P(r) = \frac{U(r)}{Total\ score} I \times N \quad (6)$$

Here $I \times N$ is the total number of virtual sample to be generated. Scaled by the normalized score of a rule, it is the number of samples to be generated from that rule.

Thus the steps of sample generation can be summarized below:

1) For each rule r in the rule set $R_o$ calculate P(r).
2) Generate P(r) number of virtual examples by first creating a template from preconditions and then set up other features randomly.

### 2.4 Dataset pruning

We can only prune a sample if we can say with sufficient *confidence* that it is not giving the right classification. Just as rules can have accuracy based on dataset. A sample can also have accuracy based on ruleset. If a sample is deemed incorrect by a rule with a high sample utility score, it is more probable that it will be misclassified. On the other hand, if a high scoring rule agrees with the sample, we can have more confidence that the sample is correct. Thus, how much confidence we have on a sample can be calculated via a voting system. We calculate the total utility score for the sample and against the sample.

Every sample is assumed to have the inherent utility score of D $\epsilon$ [0,1]. This is the use of D mentioned earlier. If the value of D is high than more negative vote is required for rejection. We suggest D to be >.51. This ensures that neither Accuracy nor Scope of a rule can fully dominate the utility value and reject an instance.

The total utility of $U_s$ values is calculated using a system similar to the union of probabilities. A sample is rejected if it has overall negative utility.

$$U_s(a) \cup U_s(a) = U_s(a) + U_s(b) - U_s(a)U_s(b) \quad (7)$$

Therefore, the value of $U_s$ never exceeds 1. And also it is based on the principle that multiple low scoring rules cannot reject a sample by over powering D.

$$V_+(i) = D \cup \bigcup_{r \in R}[Conforms(i,r) \; \& \; (i_{class} = r_{class})] U_s(r) \quad (8)$$

$$V_-(i) = \bigcup_{r \in R}[Conforms(i,r) \; \& \; (i_{class} \neq r_{class})] U_s(r) \quad (9)$$

$$Conforms(i,r) = \begin{cases} 1 \; if & \forall x \; Ant(r) \; x = i_x \\ 0 & otherwize \end{cases}$$

If $V_+(i) - V_-(i) < 0$ then remove i

## 3  Experimentation with RASCAL

This section reports the experimental results of using RASCAL to augment datasets. Two real world problems, of DNA analysis [Towell & Shavlik, 1994] were experimented[1]. We compared the performance with several popular and classic learning algorithms: standard Feed-forward Neural Network [Mitchell, 1997], C4.5 [Quinlan, 1993], k-Nearest Neighbour [Aha et al., 1991] and Support Vector Machines [Vapnik, 1998] [2]. We also compared RASCAL with the neural network based theory refine system KBANN [Towell & Shavlik, 1994].

The first dataset is the promoter recognition dataset. A promoter, which is a short DNA sequence that precedes a gene sequence, is to be distinguished from a nonpromoter. The input is a sequence of 57 nucleotides (one of A, T, G or C). The dataset has 936 instances with 236 positive and 702 negative examples. The associated ruleset had 16 rules from which 64 operational rules were generated.

The second dataset is about splice-junction determination. This is a 3 class problem; the task is to determine into which of the three categories the specified DNA sequence belongs: Exon/Intron borders (EI), Intron/Exon borders (IE) or neither. The input is also a DNA sequence with 60 nucleotides. The dataset had 1007 instances selected randomly from a population of 3190. The percentage split of the classes is 25% EI, 25% IE & 50% neither examples. The associated ruleset had 65 rules from which 260 operational rules were generated.

---

[1] The datasets are available at the UCI repository.

[2] Weka 3.6.2 (www.cs.waikato.ac.nz/ml/weka/) implementations of the learning algorithms were used in the experiments.

We conducted several tests from these datasets. The first test used the full dataset for refinement and 43% new virtual examples were added therefore I=.43 was chosen. The refined dataset and the original dataset performance was compared. We also show how the algorithms fared when given only 50% of the original dataset. The first test was conducted using 10 fold cross validation while the second test with 30% data used the remaining 70% as test set. Thus the second test much more difficult than the first. The final test was the performance of the algorithms trained only by the virtual samples. This is a curious test as this shows the knowledge virtual samples contain. The (%) accuracy rates we report are simple averages of 20 trials. The results are presented in Table 1 & 2.

| Learner | Original data-full | RASCAL Refined | Orig-data-30% | RASCAL refined 30% | Virtual Samples only |
|---|---|---|---|---|---|
| Neu.Net | 92.45 | 95.14 | 85.33 | 91.44 | 78.23 |
| SVM | 89.74 | 93.78 | 85.25 | 89.76 | 80.12 |
| C4.5 | 89.95 | 92.29 | 81.98 | 89.87 | 86.39 |
| Near. N. | 87.49 | 88.33 | 85.64 | 84.97 | 84.74 |
| KBANN | 93.70 | 95.36 | 92.21 | 92.45 | 77.54 |

*Table 1: Promoter dataset results. (% correctness)*

| Learner | Original data-full | RASCAL Refined | Orig-data-30% | RASCAL refined 30% | Virtual Samples only |
|---|---|---|---|---|---|
| Neu.Net | 91.23 | 93.69 | 83.85 | 89.48 | 76.05 |
| SVM | 88.77 | 94.92 | 85.53 | 92.34 | 81.18 |
| C4.5 | 90.86 | 93.51 | 82.83 | 90..32 | 83.00 |
| Near. N. | 80.28 | 89.26 | 67.76 | 81.09 | 75.61 |
| KBANN | 93.70 | 93.45 | 89.30 | 91.31 | 75.98 |

*Table 2: Spice-junction dataset results. (% correctness)*

It is apparent from the results that RASCAL augmented datasets outperform the original datasets. This finding is statistically significant. Unexpectedly, it even improves the performance of KBANN in both datasets which also uses the same ruleset as domain knowledge. The performance gain is even more pronounced in the test with 30% dataset. All of the algorithms show significant performance gains. The performance in the 30% dataset shows RASCAL speeds up the learning process requiring smaller amount of data which is also statistically significant. Interesting result comes from learning with virtual samples only which also shows a degree of good performance. This is the knowledge imbued in the ruleset. Our conclusion is, the performance of the virtual samples depend on the strength of the domain knowledge. Strong domain knowledge would speed up the learning process to a large extent.

The advantage of RASCAL can be more clearly evident by the learning curves in Figure 1 & 2. The test was conducted with the full dataset. The training examples were selected randomly while the rest became test set. The error rates are averaged over all the algorithms.

RASCAL augmented datasets clearly require less data. Another useful insight is the fact that, when very low data is given, RASCAL augmented datasets still perform at a good level. While using the algorithms with original data suffers from poor initial performance. As more data is provided, performance on the original data catches up.

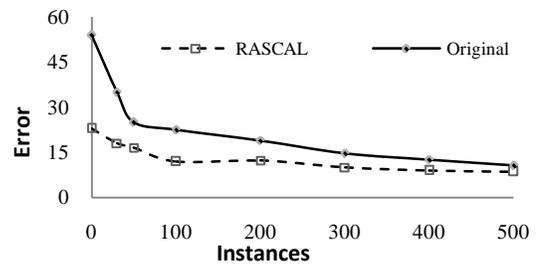

*Figure 1: Promoters Dataset Learning curve*

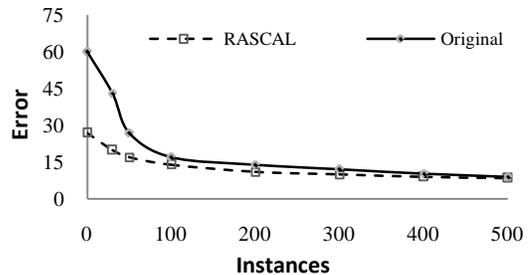

*Figure 2: Splice-junction Dataset Learning curve*

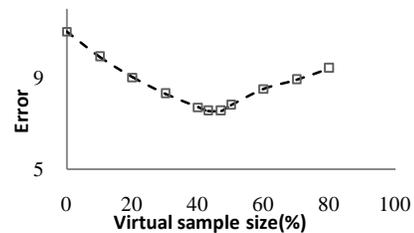

*Figure 3: Performance curve for Virtual sample size.*

An important decision to be made while learning is how much virtual samples are to be generated. Generating fewer samples will not be beneficial; while too many virtual samples will inevitably lead to over-fitting. We conducted experiments on the full Promoters dataset by gradually increasing the value of I which controls the amount of samples to be generated. The results are shown in Figure 3.

Increasing virtual samples boost performance. However, after increasing the size around 40%-50% , the performance declines and the error curves takes an upward trend. Therefore, virtual samples increase performance up to extent. After that the utility gets diminished. So, the conclusion can be made that the value I must be carefully chosen through trial.

## 4    Conclusion and Future Research

The learning process of humans is largely based augmenting knowledge previously acquired. This is the only viable way to substantially increase learning performance. Current inductive algorithms are extremely good at learning from training data. Therefore the method we proposed provides a generalized method that can be applied to increase the performance of any inductive learning algorithm. Refining the datasets enables us to use the tried and tested inductive algorithms while also having more speedy yet accurate learning. The experimental results also prove our point.

The use of virtual examples is a new field with many research possibilities. Our algorithm currently supports only propositional rules. It can be extended to learn from association rules as well. Further research could be done to enable more powerful knowledge representation techniques such as first order logic or Description logics. Even the current propositional rule based method can be extended to learn from association rules as well. Virtual sample generation can be used to incorporate other form knowledge as well. Currently no framework exists for hybrid learning algorithms. So, having a structured framework standing on firm theoretical basis is also a goal for hybrid learning research. The possibilities are almost endless.